\documentclass[10pt,twocolumn,letterpaper]{article}

\usepackage{cvpr}
\usepackage{times}
\usepackage{epsfig}
\usepackage{graphicx}
\usepackage{amsmath}
\usepackage{amssymb}
\usepackage{authblk}

\usepackage[breaklinks=true,bookmarks=false]{hyperref}

\cvprfinalcopy % *** Uncomment this line for the final submission

\def\cvprPaperID{****} % *** Enter the CVPR Paper ID here

\begin{document}

%%%%%%%%% TITLE
\title{Very Power Efficient Neural Time-of-Flight}

\author[ ]{Yan Chen$^{1,2}$\hspace{1.5mm}\thanks{Indicates equal contribution.} \quad Jimmy S. Ren$^{1\hspace{1.5mm}*}$  \quad Xuanye Cheng$^1$ \quad Keyuan Qian$^2$ \quad Jinwei Gu$^1$\vspace{-2mm}}
\affil[ ]{$^1$SenseTime Research \quad $^2$Tsinghua University}
\affil[ ]{yan-chen16@mails.tsinghua.edu.cn, qianky@sz.tsinghua.edu.cn}
\affil[ ]{\{rensijie, chengxuanye, gujinwei\}@sensetime.com}

\maketitle
%\thispagestyle{empty}

%%%%%%%%% ABSTRACT
\begin{abstract}
Time-of-Flight (ToF) cameras require active illumination to obtain depth information thus the power of illumination directly affects the performance of ToF cameras. Traditional ToF imaging algorithms is very sensitive to illumination and the depth accuracy degenerates rapidly with the power of it. Therefore, the design of a power efficient ToF camera always creates a painful dilemma for the illumination and the performance trade-off. In this paper, we show that despite the weak signals in many areas under extreme short exposure setting, these signals as a whole can be well utilized through a learning process which directly translates the weak and noisy ToF camera raw to depth map. This creates an opportunity to tackle the aforementioned dilemma and make a very power efficient ToF camera possible. To enable the learning, we collect a comprehensive dataset under a variety of scenes and photographic conditions by a specialized ToF camera. Experiments show that our method is able to robustly process ToF camera raw with the exposure time of one order of magnitude shorter than that used in conventional ToF cameras. In addition to evaluating our approach both quantitatively and qualitatively, we also discuss its implication to designing the next generation power efficient ToF cameras. We will make our dataset and code publicly available.
\end{abstract}

%%%%%%%%% BODY TEXT
\section{Introduction}
\begin{figure}[!t]
\centering
\footnotesize
\renewcommand{\tabcolsep}{1pt} % adjust horizontal space
\renewcommand{\arraystretch}{1} % adjust vertical space
\begin{center}
\begin{tabular}{cc}
  \includegraphics[width=0.49\linewidth]{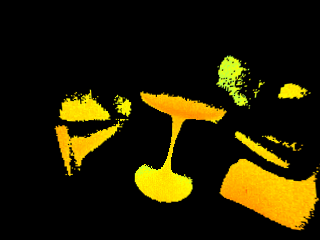} &
  \includegraphics[width=0.49\linewidth]{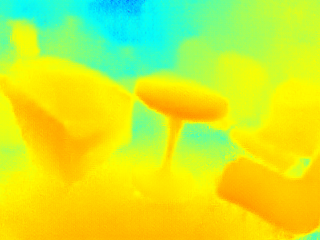} \\
  (a) Conventional depth map under&(b) Our result from\\
  extreme short exposure&ToF raw of (a)\\
  \includegraphics[width=0.49\linewidth]{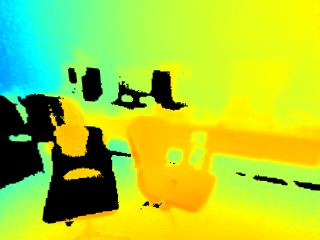} &
  \includegraphics[width=0.49\linewidth]{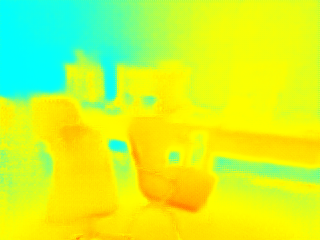} \\
  (c) Conventional depth map under&(d) Our result from\\
  regular exposure&ToF raw of (c)
\end{tabular}
\end{center}
\vspace{-2mm}
\label{fig:figure1}
\caption{We propose an end-to-end pipeline to translates the weak and noisy ToF camera raw to high quality depth map. (a) Depth image produced by the ToF camera's default imaging pipeline with 200us exposure time. The quality is very poor. (b) Depth image produced by our method applied to the ToF camera raw from (a). (c) Depth image produced by the ToF camera's default imaging pipeline with regular exposure time. Some of the depth information is still lost due to objects with low reflectivity or long distances. (d) Depth image produced by our method applied to the ToF camera raw from (c).}
\vspace{-5mm}
\end{figure}
Depth sensing is one of the core components of many computer vision tasks. Amplitude-modulated continuous-wave (AMCW) time-of-flight (ToF) has a brief and definite physical meaning in depth construction of scenes thus it attracts a lot of commercial attention, such as Kinect V2. It is also widely used in academic research of computer vision \cite{Alpher01, Alpher02}, including human tracking \cite{Alpher03}, 3D scene reconstruction \cite{Alpher04}, robotics \cite{Alpher05}, object detection, gesture recognition \cite{Alpher06, Alpher07}, and scene understanding \cite{Alpher08, Alpher09}. However, comparing with traditional RGB cameras, ToF cameras compute the depth by emitting a periodic amplitude modulated illumination signal and receive the demodulated signal reflected by the objects. Higher power of active illumination enables the ToF sensor to receive the signal with higher signal noise ratio (SNR) and higher level of confidence. Therefore, the power of illumination directly influences the performance of ToF cameras.

Traditional ToF imaging algorithms are very sensitive to illumination and the depth accuracy degenerates rapidly with the decreasing illumination power. In order to obtain more accurate depth information, one way is to increase the intensity of the received active illumination signal. Other than increasing the illumination power, an alternative treatment to this issue is to increase the physical size of the pixels on the sensor to collect more light. However, this significantly decreases the depth map resolution. According to the inverse square law, one can also cut the depth sensing range of the camera. This obviously decreases the usability of the camera in many applications. Therefore, to make a ToF camera with satisfactory depth quality as well as reasonable resolution and sensing range, the painful dilemma for the illumination and the performance trade-off always troubles the designer of the camera if the conventional imaging pipeline is used.  

Such dilemma can be tackled if there is a way to recover high quality depth information from weak signals. A number of recent studies show that it is plausible to recover high SNR natural images from very noisy data using deep learning. \cite{Alpher10, Alpher11, Alpher12}. Chen et al. \cite{Alpher12} showed impressive results on recovering high quality color image from camera Bayer pattern which is captured under extremely low light condition with short exposure. Inspired by these research, we show for the first time that for ToF cameras, despite the weak signals in many areas under the extreme short exposure setting, these signals as a whole can be well utilized through a learning process which directly translates the weak and noisy ToF camera raw to high quality depth map. This creates an opportunity to address the aforementioned dilemma and makes it possible to design a very power efficient ToF camera possibly with higher resolution and longer sensing range. To enable the learning, we collect a comprehensive dataset under a variety of scenes and photographic conditions via a specialized ToF camera. The dataset contains ToF raw measurements and depth maps collected under extreme short exposure settings and long exposure settings respectively. We show in the experiments that our proposed method is able to robustly process ToF raw measurements with an exposure time that is one order of magnitude shorter than that used in a conventional ToF camera.

The contributions of our work can be summarized as follows. 
\begin{itemize}
\item We show for the first time that our proposed method is able to recover high quality depth information from very weak ToF raw data (one order of magnitude shorter exposure time).
\item We introduce a real-world dataset used for training and validating the this learning tasks. We will make the code and dataset publicly available.
\item We shed light on the design of the next generation ToF camera by providing an effective alternative to optimize the performance and power consumption trade-off.
\end{itemize}

\section{Related Work}
\noindent{\bf Depth reconstruction based on ToF cameras.} ToF cameras face a lot of challenging problems when extracting depth from raw phase-shifted measurements with respect to emitted modulated infrared signal. Dorrington et al. \cite{Alpher13} established a two-component, dual-frequency approach to resolving phase ambiguity, achieving significant improvements of the accuracy when distortion is caused by multipath interference (MPI). Several methods were proposed to deal with MPI distortions, including adding or modifying hardware \cite{Alpher14,Alpher15,Alpher16}, employing multiple modulation frequencies \cite{Alpher13,Alpher17,Alpher18,Alpher19} and estimating light transport through an approximation of depth \cite{Alpher20,Alpher21}. Marco et al. \cite{Alpher22} correct MPI errors by a two-stage training strategy, training the encoder to represent MPI-corrupted depth images with captured dataset firstly and then use synthetic scenes to train the decoder to correct the depth. However, the above pipelines are based on the assumption that there is no cumulative error and information loss introduced in the previous stage, thus the final result of these methods is likely to contain cumulative errors of multiple stages.

Krishna et al. \cite{Alpher24} filled the missing depth pixels by using a color-aware Gaussian-weighted averaging filter to estimate depth value. However, its performance is limited by the similarity between the neighborhood pixels and target pixels and the information of the target region is wasted. An end-to-end ToF image processing framework presented by Su et al. \cite{Alpher23} can efficiently reduce noise, correct MPI and resolve phase ambiguity. However, the training data is not realistic. Therefore, depth reconstruction may fail when the scene contains low reflectivity materials and objects. To the best of our knowledge, none of existing depth reconstruction method is able to obtain high quality depth map from the weak and noisy ToF camera raw measurements.\\\vspace{-1mm}

\noindent{\bf Image enhancement under low light.} For conventional RGB cameras, photography in low light is challenging. Several techniques have been proposed to increase the SNR of the recovered image \cite{Alpher25,Alpher26,Alpher27,Alpher28,Alpher30}. Chen at el.\cite{Alpher12} established a pipeline by training a fully convolutional neural network which directly translate the very noise and dark Bayer pattern camera raw to high quality color images. Though impressive results from the aforementioned studies, deep learning and data-driven approaches have not yet been adopted to recover high quality depth information from weak and noisy ToF raw. It remains unclear if such methodology is effective for ToF imaging. The aim of this paper is to disclose its feasibility.\\\vspace{-1mm}

\noindent{\bf Depth datasets.} Although recently many datasets of depth maps are proposed, most of them are consisted of synthetic data, such as transient images generated via time-resolved rendering. A dataset of ToF measurements \cite{Alpher22} is proposed via simulating 25 different scenes with a physically-based, time-resolved renderer. Besides, Su et al. \cite{Alpher23} offer a large-scale synthetic dataset of raw correlation time-of-flight with ground truth labels. However, the ToF raw with artificial distortions and Gaussian noise is not realistic enough to support the real life generalization especially when dealing with areas with large noise caused by low reflectivity. Only the raw RGB data, depth map and accelerometer data are provided in the NYU-Depth V2 dataset \cite{Alpher35} but ToF raw measurements are missing. Thus, this dataset can not be used to train ToF raw to depth map conversion. Furthermore, most existing depth datasets concentrate on images captured under appropriate illumination or ideal environments, they are not suitable for evaluating imaging with low active illumination power or weak reflected signal. In this paper, we propose a comprehensive dataset to fill these gaps and enable the training and validation of our proposed model.

\section{Method and Analysis}
\subsection{Imaging with Time-of-Flight Sensors}
\noindent{\bf Distance measurement.} The distance measurement mode of Time of Flight uses the on chip driver and the external LED/LD to provide modulated light on the target. Generally, the period of the modulation control signal is programmable. The modulator generates all signals to modulate the external LED/LD and simultaneously all demodulation signals to the pixel-field. We can describe the programmable modulation optical signal with angular frequency $\omega$ as
\begin{equation}
s(t) = \cos(\omega t),
\end{equation}
where the amplitude is normalized. Once the signal is reflected by the object, the modulated optical signal goes back to the sensor with certain amplitude attenuation and certain phase shift, then the received signal can be expressed as
\begin{equation}
r(t) =  \alpha \cos(\omega t-\varphi) + \delta,
\end{equation}
where $\delta$ is the offset, $\alpha$ is the amplitude after attenuation, and $\varphi$ is the phase shift. In order to achieve demodulation, the original emission signal needs to be used as a correlation signal and demodulated with the received signal as
\begin{equation}
\begin{split}
\varphi _{sr}&= r(t) \otimes s(t) \\
&= \lim_{T \to \infty} \frac{1}{T}\int_{-T/2}^{T/2} s(t)g(t + \tau)\, dt \\
&= \lim_{T \to \infty} \frac{1}{T}\int_{-T/2}^{T/2} [ \alpha \cos(\omega t-\varphi)+\delta ][\cos(\omega t + \omega \tau)]\, dt \\
&= \frac{\alpha}{2}\cos(\omega t+\varphi),
\end{split}
\end{equation}
where the relevant signal is denoted as $C(\tau) = \varphi _{sg}(\tau)$. ToF cameras need to sample the correlation signal $C(\tau)$ four times in one cycle. That is, sampling is performed when $\omega \tau _0 = 0^\circ, \omega \tau _{90} = 90^\circ, \omega \tau _{180} = 180^\circ, \omega \tau _{270} = 270^\circ$. Considering the received signal is mainly superimposed on the background image, we also need to consider an offset here. Then the phase shift $\varphi$ and the amplitude $\frac{\alpha}{2}$ can be obtained from the four sample values
\begin{equation}
\varphi = \arctan\frac{C(\tau _3) - C(\tau _1)}{C(\tau _0) - C(\tau _2)},
\end{equation}
\begin{equation}\label{equ:Eq5}
\alpha = \frac{\sqrt{[C(\tau _3) - C(\tau _1)]^2+[C(\tau _0) - C(\tau _2)]^2}}{2},
\end{equation}
and we can find the distance value by the phase shift
\begin{equation}
d = \frac{c\varphi}{2\omega},
\end{equation}
where c is the speed of light.\\\vspace{-1mm}
\begin{figure}[!t]
\centering
\includegraphics[scale=0.25]{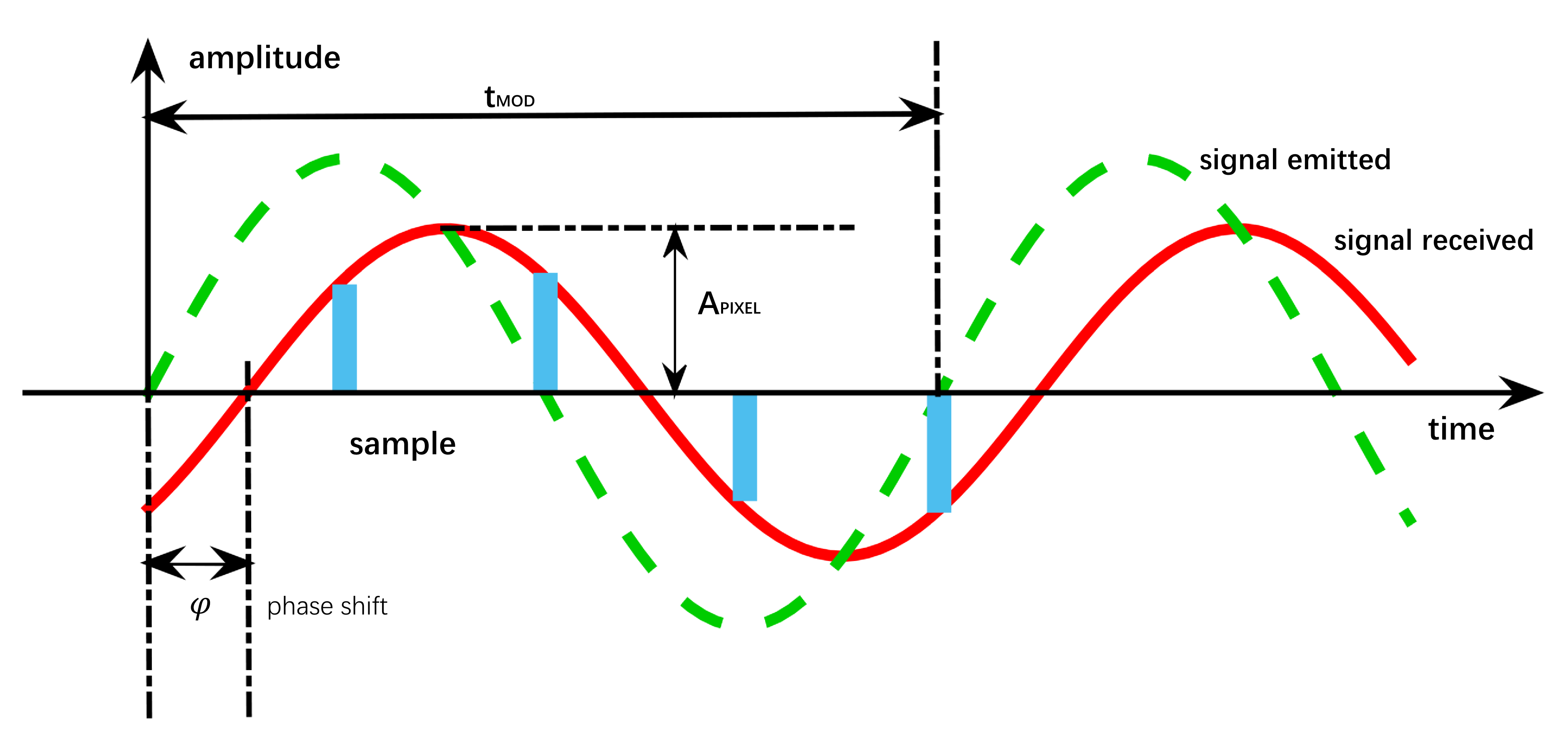}
\caption{Sample of received signal per $\pi$/4.}
\label{fig:figure2}
\vspace{-3mm}
\end{figure}
\begin{table*}[]
\begin{center}
\resizebox{0.98\linewidth}{!}{
\begin{tabular}{cccccccccc}
\hline
\multicolumn{10}{c}{Network Architecture} \\ \hline
Name &D1& D2 & D3 & D4 & Res1-Res9 & U1 & U2 & U3 & U4 \\ \hline
Layer & \begin{tabular}[c]{@{}c@{}}conv+\\ LeakyReLU\end{tabular} & \begin{tabular}[c]{@{}c@{}}conv+\\ LeakyReLU\end{tabular} & \begin{tabular}[c]{@{}c@{}}conv+\\ LeakyReLU\end{tabular} & \begin{tabular}[c]{@{}c@{}}conv+\\ LeakyReLU\end{tabular} & ResBlock & \begin{tabular}[c]{@{}c@{}}deconv\\ +ReLU\end{tabular} & \begin{tabular}[c]{@{}c@{}}deconv\\ +ReLU\end{tabular} & \begin{tabular}[c]{@{}c@{}}deconv\\ +ReLU\end{tabular} & \begin{tabular}[c]{@{}c@{}}deconv\\ +Tanh\end{tabular} \\ \hline
Kernel & 4$\times$4 & 4$\times$4 & 4$\times$4 & 4$\times$4 & 3$\times$3 & 4$\times$4 & 4$\times$4 & 4$\times$4 & 4$\times$4 \\ \hline
Stride & 2 & 2 & 2 & 2 & 1 & 2 & 2 & 2 & 2 \\ \hline
I/O & 4/64 & 64/128 & 128/256 & 256/256 & 256/256 & 256/256 & 512/128 & 256/64 & 128/3 \\ \hline
Input & ToF raw & D1 & D2 & D3 & D4 & Res9 & D3+U1 & D2+U2 & D1+U3 \\ \hline
\end{tabular}
}
\caption{Architecture of our network. "conv" represents a convolutional layer. "deconv" means a fractionally-strided convolutional layer. We use leaky ReLUs with a negative slope of 0.2. "ResBlock" means a residual block that contains two 3$\times$3 convolutional layer with the same number of filters and a ReLU layer is in the middle.}
\label{network table}
\end{center}
\vspace{-5mm}
\end{table*}

\noindent{\bf Quality of the measurement result.} Raw ToF measurements contain the distance information, as well as the quality and the validity (confidence level) of the received optical signal. A higher amplitude of the measured signal represents a more accurate distance measurement. The depth data for each pixel has its own validity and quality in ToF cameras. The amplitude of the modulated light received by the ToF sensor is the primary quality indicator for the measured distance data. It can be calculated as Eq.\ref{equ:Eq5}. However, excessive active illumination will make the amplitude of the raw measurements very large. This leads to errors in the depth value due to the problem of over-exposure of the ToF sensor.\\\vspace{-1mm}

\noindent{\bf Problems of Traditional Pipeline.} In order to recover high-quality depth maps from imperfect ToF raw measurements, traditional methods of ToF camera imaging often require a series of specialized processing techniques, such as denoising, correction of multipath distortion and nonlinear compensation, etc. However, these components are independent to each other and often relies on the assumption of no cumulative error and information loss in the previous stages. In practice, this assumption is almost always not true. It may cause large errors in the final depth map. To alleviate the overall error, a distance calibration process is conducted to adjust the offset value to the selected calibration plane and sets the Fix Pattern Noise (FPN) on the plane to zero. However, this technique can not be generalized to scenarios of weak signals. 

As mentioned above, the amplitude of the modulated signal received by the ToF sensor is the primary quality indicator for the measured distance data. When the amplitude is lower than a certain threshold, the traditional ToF imaging method is unable to calculate a reliable depth value at such a low SNR, so that the depth information is missing in these areas (behave as a black hole on the depth map). The experiment results show that the condition for invalidating the traditional ToF camera imaging pipeline in a pixel as:
\begin{equation}
A_{pixel} < 0.024A_{max} ,
\end{equation}
where $A_{max}$ is the maximum amplitude that can be imaged by the ToF sensor chip.
\subsection{Learning from imperfect ToF camera raw}
In this section, our approach of depth reconstruction is presented in detail. We first describe the advantage of our method of recovering high-quality depth images from weak and noisy ToF camera raw measurements compared to traditional ToF imaging methods. Then, we give a brief description of our whole pipeline to learn a mapping from ToF measurements acquired under low power illumination to corresponding high-quality depth map. And the network architecture of our method, as shown in Tab.\ref{network table}, will be introduced. Finally, we present how we train the model and implementing details.\\\vspace{-1mm}

\noindent{\bf Comparison to traditional pipeline.} The raw ToF measurements have a very low signal-to-noise ratio (SNR) and amplitude intensity, when the active illumination signal received by the ToF sensor is very low. In this case, conventional edge aware filtering methods such as bilateral filter tend to fail. Traditional method of ToF measurements denoising is based on arbitrary rules and assumptions, but these rules and assumptions often become invalid with changes in scenes and intensity of the received signals. This is particularly true for weak input signals. Therefore, it is very difficult to select the optimal parameters for all the image processing components to achieve good results for all scenarios. In contrast, the proposed method adopts the end-to-end learning and inference approach to translate the weak and noisy ToF camera raw to high quality depth map which avoids the highly complex parameter tuning for such noisy and weak input signals.\\\vspace{-1mm}
\begin{figure*}[!t]
\centering
\footnotesize
\renewcommand{\tabcolsep}{3pt} % adjust horizontal space
\renewcommand{\arraystretch}{1} % adjust vertical space
\begin{center}
\begin{tabular}{cccc}
  \includegraphics[width=0.275\linewidth]{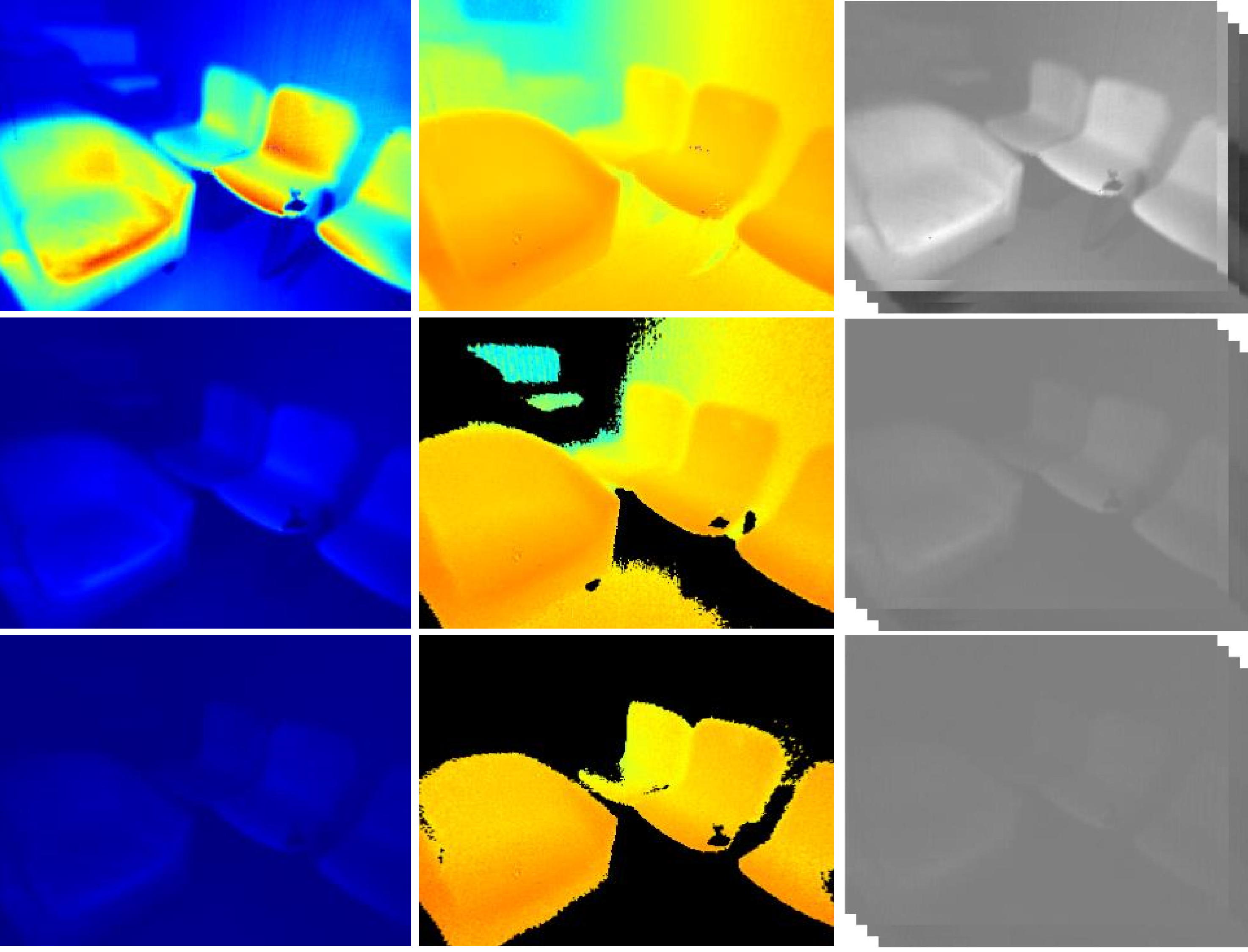} &
  \includegraphics[width=0.275\linewidth]{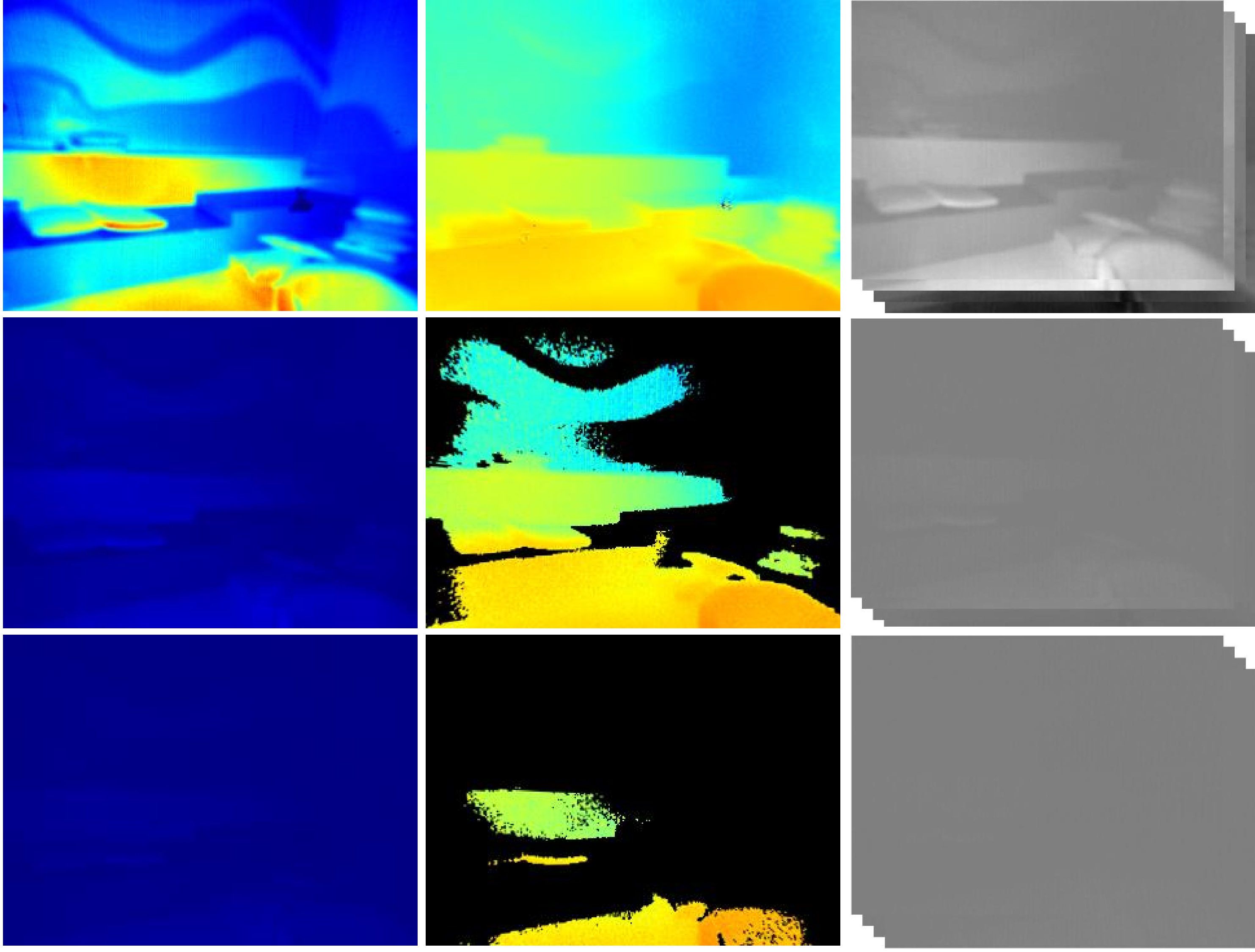} &
  \includegraphics[width=0.255\linewidth]{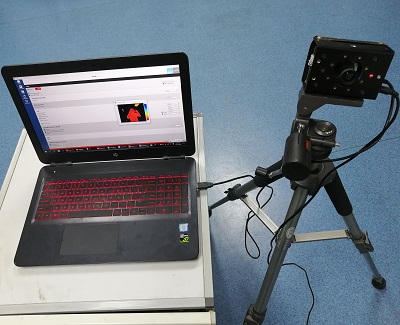} &
  \includegraphics[width=0.143\linewidth]{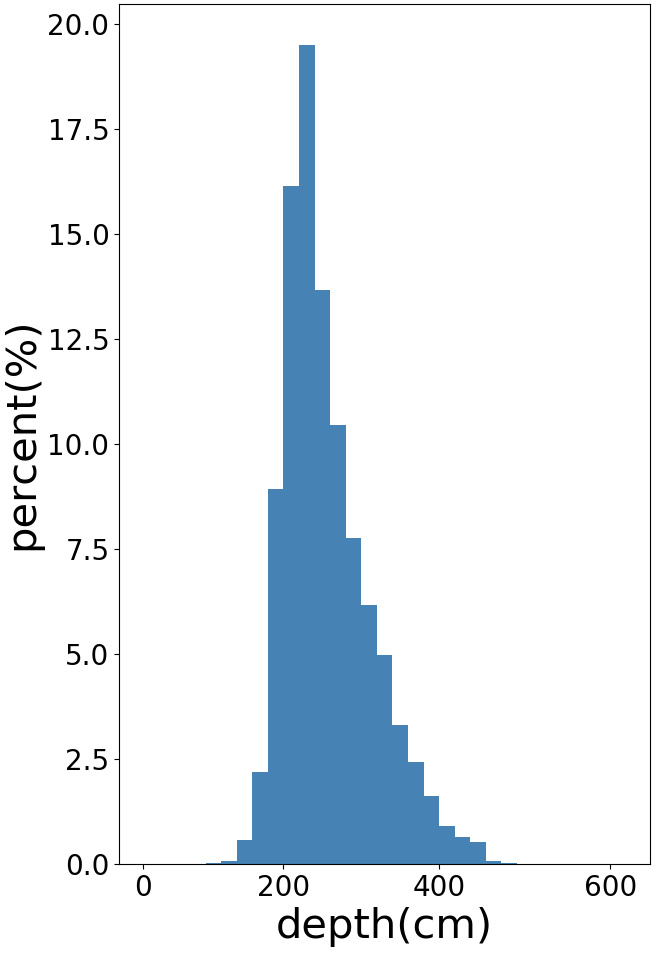} \\
  (a) a group of measurements&(b) a group of measurements&ToF camera&Depth statistics.
\end{tabular}
\end{center}
\vspace{-2mm}
\caption{We use EPC660 ToF camera from ESPROS to collect a dataset of multiple pairs of short-exposure and corresponding long-exposure depth measurements. Diverse indoor scenes are collected in the dataset, including office room, restaurant, bedroom and Laboratory. The depth range is reasonable for indoor scenes of our dataset, for depth values range from 0cm to 591cm and has a mean of 236.88cm.}
\label{fig:figure4}
\vspace{-2mm}
\end{figure*}

\noindent{\bf Our pipeline.} %Given the ToF raw measurements, our end-to-end network architecture can generate high-quality depth map even there are some weak and noisy signals caused by large distance or objects with high absorptivity in the scene. The raw ToF measurements contain more depth information and detailed structures, compared with the depth map produced by traditional ToF method, when the captured signals are weak and noisy. Therefore, the ToF raw measurements captured under short exposure are taken as the four-channel input of our method and the depth map captured under regular exposure are regarded as the ground truth for network training.
To build intuition for this end-to-end approach, we have analyzed several previous work of image-to-image mapping. Most of them have adopted an encoder-decoder network with or without skip connections \cite{Alpher32}, which are consisted of down-sampling, residual blocks and up-sampling. The pixel value of ToF depth map is closely-related to camera settings, scene architecture and layout, compared with RGB images. Besides, the geometry and architecture of scene for both depth map and raw measurements are required to be consistent. And these specific characteristics of ToF raw measurements should be combined with the previous work of image translation, when designing network architecture. 

For the above considerations, we select the encoder-decoder with skip connections as our network architecture. The size of input is progressively decreasing in pace with going through the down-sampling layers for four times, until it reaches the residual blocks. And after passing through nine residual blocks and four up-sampling layers, the size of input becomes larger and restored to its original size. The strided convolution layers combined with activation layers serve as decoder and the fractional convolution layers combined with activation layers are regarded as encoder. The residual blocks without normalization are adopted by the bottle neck part. Moreover, we added the skip connections to the network between each pair of layer i and layer n-i following the U-net to enhance the accurate of results.

To obtain high-quality depth reconstruction results, L1 loss is adopted to train our network:
\begin{equation}\label{equ:Eq8}
L_1 =\frac{1}{N}\sum_{i,j}^{N}\left| d^{gt}_{i,j}-d^{pre}_{i,j} \right| ,
\end{equation}

\noindent{\bf Training details.} Our networks is implemented in Pytorch. During training, inputs of the network are the ToF raw measurements captured under short exposure and the ground truth is the corresponding depth map captured under regular or long exposure. We randomly crop out 128$\times$128 images on the original 320$\times$240 images for data augmentation. This strategy effectively improves the robustness of the model. We train our network using the Adam optimizer \cite{Alpher40} with an initial learning rate of 0.0002 for the first 200 epochs, before linearly decaying it to 0 over another 1800 epochs. 

\section{Dataset}
To enable the learning, we collect a comprehensive dataset under a variety of scenes and photographic conditions by a specialized ToF camera with raw data access. Due to the limitations of hardware devices, it is difficult to change the intensity of received signals by directly changing the physical size of the pixels on the ToF sensor or the power of the infrared LED illumination of the development kit. However, we can modify the intensity of received signal by changing the exposure time of the ToF camera, since the exposure time is directly proportional to the intensity of received signal.

We use EPC660 ToF camera from ESPROS to collect a dataset of multiple pairs of short-exposure and corresponding long-exposure depth measurements for training the proposed architecture. ToF raw measurements, amplitude image and depth map at 320$\times$240 resolution are collected for each scene with an exposure time. We captured 200 groups of measurements with 200us and 400us exposure time respectively and 200 groups of corresponding long-exposure images from a variety of scenes with varying materials. During the experiments, we use 150 groups for training and 50 groups for testing.

Diverse indoor scenes are collected in the dataset, including office room, restaurant, bedroom and laboratory. We adopt the ideal sinusoidal modulation functions to avoid the wiggling effect. The images are generally captured at night in rooms without infrared monitoring to avoid the influence of solar radiation and infrared light emitted by some particular machines. Note that a variety of hard cases such as distant objects, fine structures, irregular shapes and various materials including fabric, metals with low reflectivity and dark object with high absorptivity exist in our scenes.

We mount the ToF camera on a sturdy tripod to avoid camera shaking and other vibration when capturing. Due to continuous modulation, 6MHz was selected as modulation frequency for measuring depth in our scenes with range of 0-6 meters to prevent roll-over being observed. Then exposure time is adjusted to obtain high-quality raw data. After long-exposure ToF measurements captured, we decrease the exposure time to 200 us and 400 us respectively via software on computers to collect data without touching the cameras.
\begin{table}[!t]
\begin{center}
\resizebox{0.98\linewidth}{!}{
\begin{tabular}{ccccc}
\hline
                                                                & \multicolumn{2}{c}{200us} & \multicolumn{2}{c}{400us} \\ \hline
                                                                & MAE     & SSIM    & MAE     & SSIM   \\ \hline
\begin{tabular}[c]{@{}c@{}}Traditional \\ pipeline\end{tabular} & 179.79      & 0.1615      & 93.39       & 0.5162      \\ \hline
Ours                                                            & 10.13       & 0.9156      & 7.94        & 0.9342      \\ \hline
\end{tabular}
}
\end{center}
\vspace{-2mm}
\caption{This table reports the mean absolute error (MAE)(cm) and the structural similarity (SSIM)(\%) of 200us exposure time and 400us exposure time. Traditional method can recover depth information only in the local position under the low exposure setting, so the overall error is very large.}
\label{table1}
\vspace{-3mm}
\end{table}
A mask to evaluate the quality of ToF measurements will be introduced into our dataset. Actually, the quality and validity of the received signal exists in raw data collected by ToF cameras. The signal amplitude as well as the ratio of ambient-light $E_{BW}$ to the value of modulated light $E_{ToF}$ (AMR) indicates the quality and validity of received signal. We combine these two features of received signals in a certain proportion to generate a quantitative criteria for evaluating the quality of each pixel in measurements. A threshold for criteria can be defined to produce a mask for each pixel. This mask can be adopted in network training and depth map generation. For instance, unconfident pixels in the labels can be ignored during the computation of error gradients in training.

Fig.\ref{fig:figure4} shows quantitative analysis of depth-range distribution of ToF measurements in our dataset. The depth range is reasonable for indoor scenes of our dataset, for depth values range from 0cm to 591cm and has a mean of 236.88cm. There are some regions with no depth value or much noise when short-exposure, due to few reflected photons detected. The ToF measurements is sufficient to serve as ground truth, though some noise still exists.

\section{Experiments and Results}
\subsection{Qualitative results}
We first quantify depth error with the mean absolute error (MAE) and the structural similarity (SSIM) \cite{Alpher41} of predicted depth map compared to the ground truth. At the same time, we will analyze the impact of different network structures on our results. Finally, we will quantitatively analyze the variation of the error of our method at different detection distances.\\\vspace{-1mm}

\noindent{\bf Effect of exposure time.} Our dataset contains raw data acquired under 200us and 400us exposure time and their corresponding depth maps collected under regular exposure time. We have trained two models on the ToF raw measurements under 200us and 400us exposure respectively, and tested the accuracy of the two models with the corresponding test set.  Then we calculate the mean absolute error (MAE) and the structural similarity (SSIM) \cite{Alpher41} and compare the results of the traditional ToF camera pipeline with that of our proposed method on the test set. Note that the result is calculated over the whole test database in which the object distance varies between 0 to 591cm as indicated in the previous section.

As shown in Tab.\ref{table1}, our results meet an overall 7.94cm depth error with raw captured under 400us exposure time and 10.13cm with raw captured under 200us exposure time. Although the accuracy of depth map produced by our method decreases with the reduction of exposure time, the experimental results of the two models both greatly exceed that of traditional pipeline method.\\\vspace{-1mm}
\begin{table}[!t]
\begin{center}
\resizebox{0.98\linewidth}{!}{
\begin{tabular}{ccccc}
\hline
                                                                & \multicolumn{2}{c}{200us} & \multicolumn{2}{c}{400us} \\ \hline
                                                                & MAE     & SSIM    & MAE     & SSIM   \\ \hline
LSGAN & 12.00      & 0.8938     & 9.03       & 0.9149      \\ \hline
U-net & 10.13       & 0.9156      & 7.94        & 0.9342      \\ \hline
\end{tabular}
}
\end{center}
\vspace{-2mm}
\caption{This table reports the mean absolute error (MAE)(cm) and the structural similarity (SSIM)(\%) of replacing the U-net \cite{Alpher32} (our default architecture) by the the the least square GAN. We can see depth map produced by the U-net have higher SSIM and lower MAE.}
\label{table2}
\vspace{-2mm}
\end{table}

\noindent{\bf Effect of network structures.} The network structure of the the least square GAN \cite{Alpher42} is used to recover depth map from ToF raw measurements by \cite{Alpher23}, we also compare the impact of applying this structure to our framework on the results. Tab.\ref{table2} reports the result of replacing the U-net \cite{Alpher32} (our default architecture) by the the least square GAN \cite{Alpher42}. The results prove that although the network structure of the the least square GAN \cite{Alpher42} has achieved great success in image transfer, in the task of recover depth map from ToF raw measurements, depth map produced by the U-net have higher SSIM and MAE.
\begin{figure}[!h]
\centering
  \includegraphics[width=0.99\linewidth]{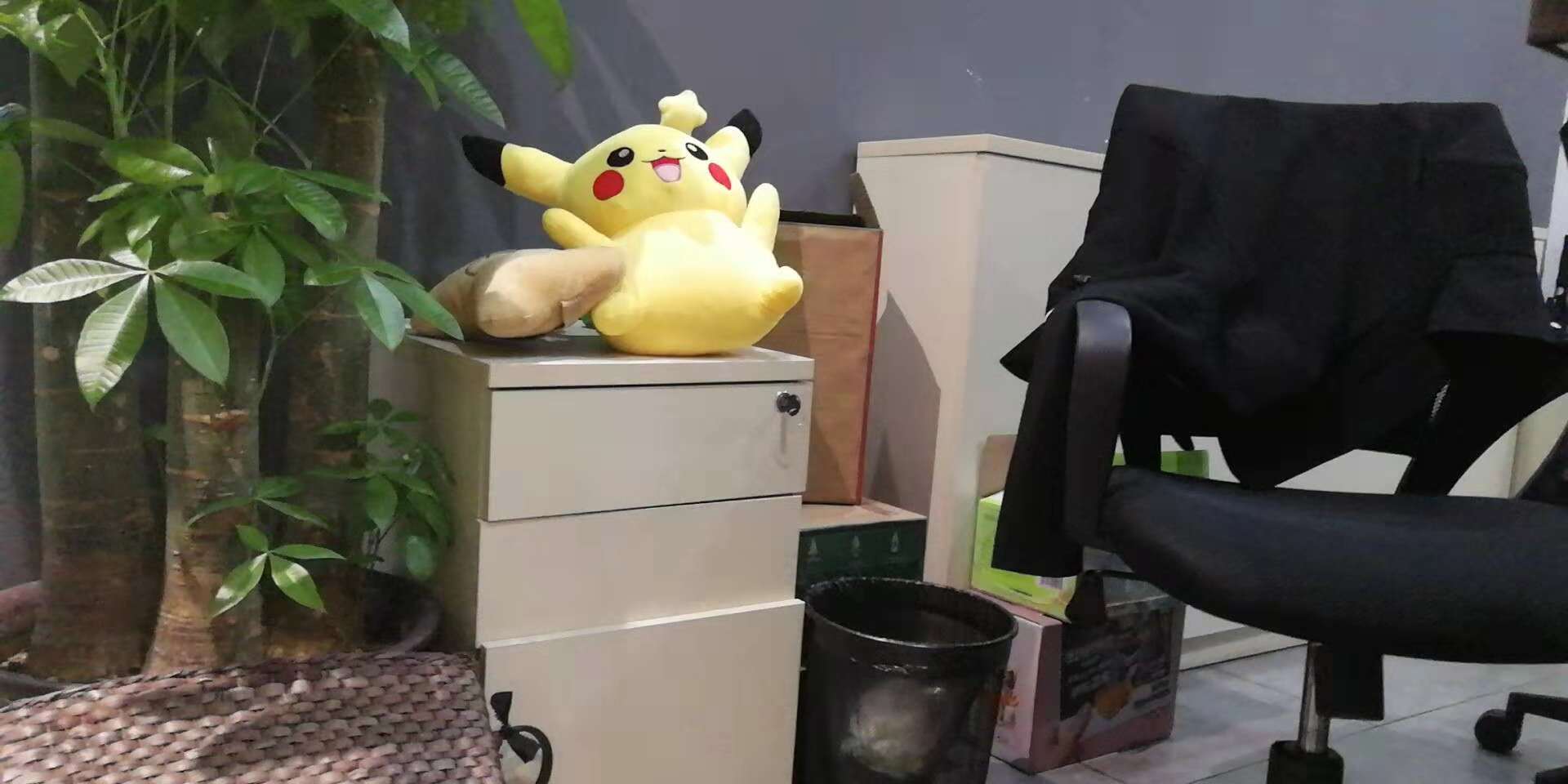}
\caption{The complex scene is designed for evaluating the performance of our method under different distances. The distance between ToF camera and the object in scene is distributed from 100cm to 200cm.}
\label{fig:figure6}
\end{figure}

\begin{figure*}[!t]
\footnotesize
\renewcommand{\tabcolsep}{1pt} % adjust horizontal space
\renewcommand{\arraystretch}{1} % adjust vertical space
\begin{center}
\begin{tabular}{cccccc}
  &200us&200us&400us&400us&4000us\\
  \includegraphics[width=0.16\linewidth]{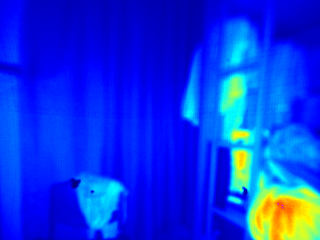} &
  \includegraphics[width=0.16\linewidth]{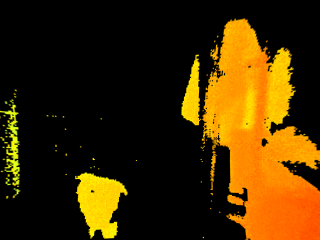} &
  \includegraphics[width=0.16\linewidth]{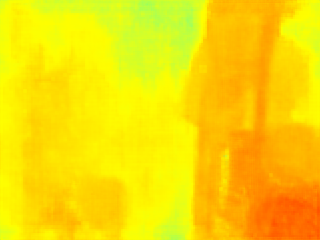} &
  \includegraphics[width=0.16\linewidth]{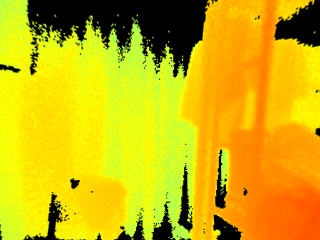} &
  \includegraphics[width=0.16\linewidth]{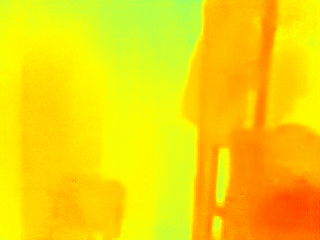} &
  \includegraphics[width=0.16\linewidth]{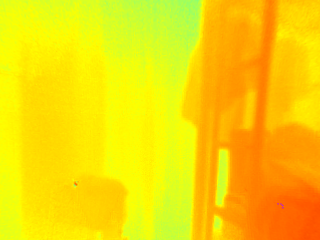} \\
  \includegraphics[width=0.16\linewidth]{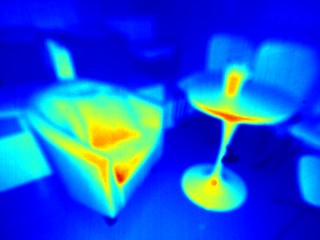} &
  \includegraphics[width=0.16\linewidth]{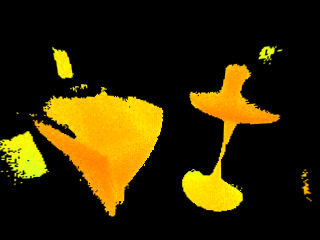} &
  \includegraphics[width=0.16\linewidth]{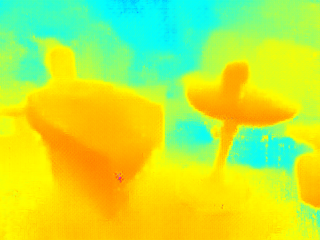} &
  \includegraphics[width=0.16\linewidth]{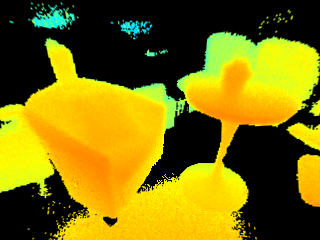} &
  \includegraphics[width=0.16\linewidth]{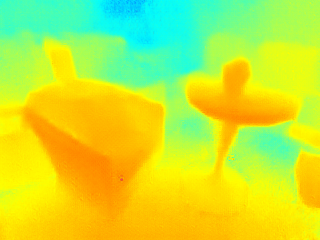} &
  \includegraphics[width=0.16\linewidth]{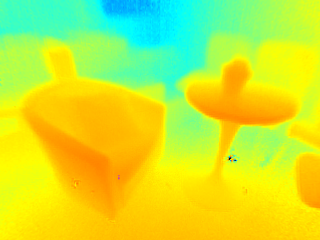} \\
  \includegraphics[width=0.16\linewidth]{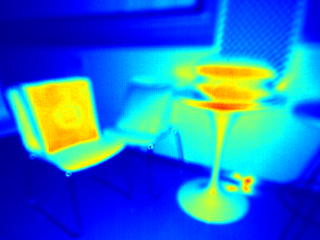} &
  \includegraphics[width=0.16\linewidth]{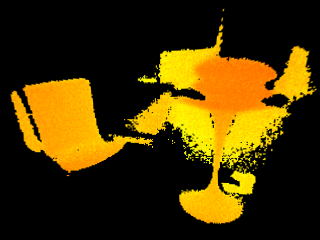} &
  \includegraphics[width=0.16\linewidth]{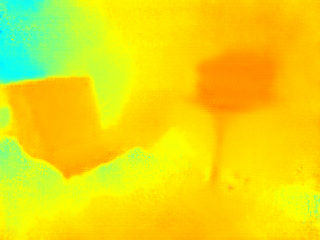} &
  \includegraphics[width=0.16\linewidth]{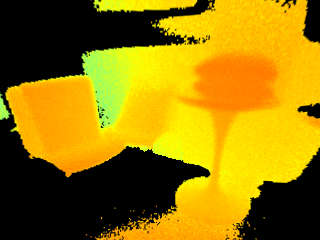} &
  \includegraphics[width=0.16\linewidth]{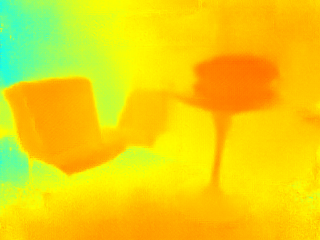} &
  \includegraphics[width=0.16\linewidth]{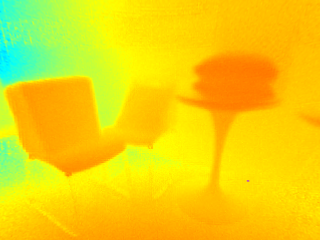} \\
  \includegraphics[width=0.16\linewidth]{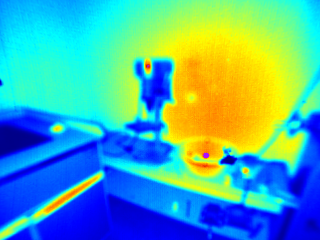} &
  \includegraphics[width=0.16\linewidth]{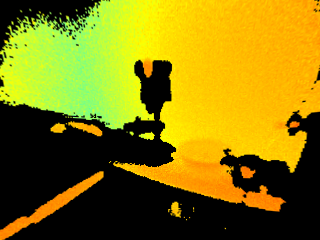} &
  \includegraphics[width=0.16\linewidth]{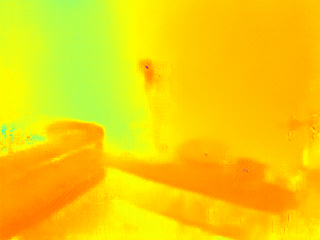} &
  \includegraphics[width=0.16\linewidth]{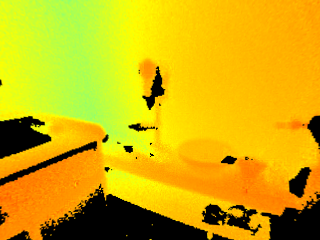} &
  \includegraphics[width=0.16\linewidth]{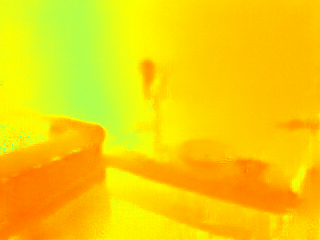} &
  \includegraphics[width=0.16\linewidth]{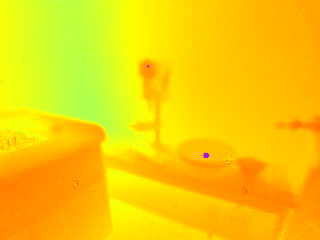} \\
  \includegraphics[width=0.16\linewidth]{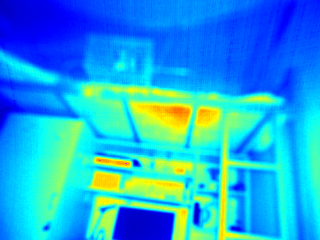} &
  \includegraphics[width=0.16\linewidth]{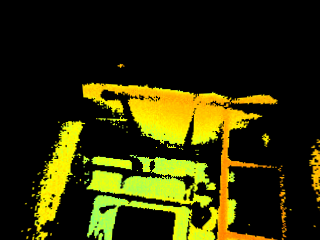} &
  \includegraphics[width=0.16\linewidth]{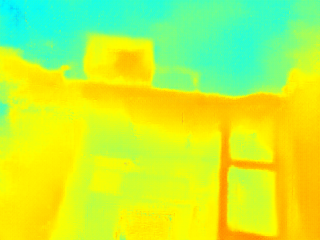} &
  \includegraphics[width=0.16\linewidth]{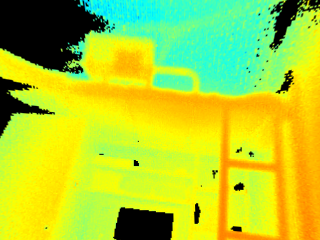} &
  \includegraphics[width=0.16\linewidth]{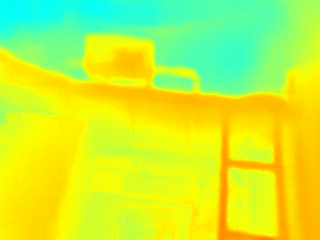} &
  \includegraphics[width=0.16\linewidth]{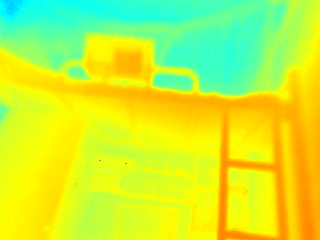} \\
  \includegraphics[width=0.16\linewidth]{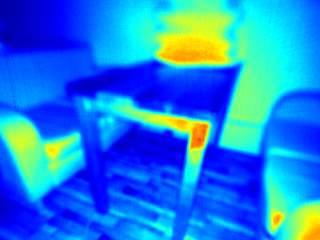} & 
  \includegraphics[width=0.16\linewidth]{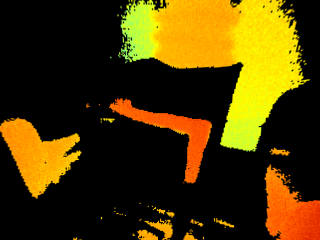} &
  \includegraphics[width=0.16\linewidth]{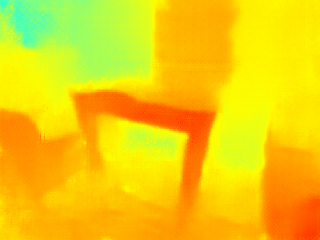} &
  \includegraphics[width=0.16\linewidth]{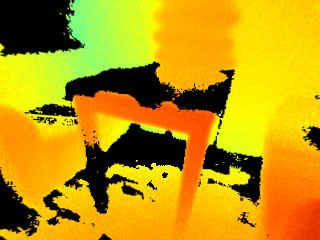} &
  \includegraphics[width=0.16\linewidth]{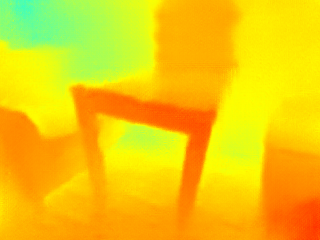} &
  \includegraphics[width=0.16\linewidth]{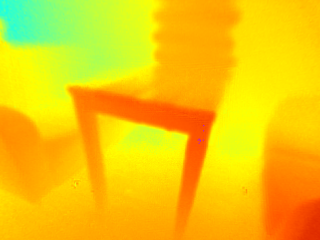} \\
  amplitude&traditional pipeline&ours&traditional pipeline&ours&ground truth
\end{tabular}
\end{center}
\vspace{-2mm}
\caption{Experiment results. In order to verify the effectiveness of our proposed method, we validated our method on our test set for exposure times of 200us and 400us respectively. The results show that our method is able to robustly process ToF camera raw with the exposure time of one order of magnitude shorter than that of conventional ToF cameras.}
\label{fig:big_figure}
\vspace{-4mm}
\end{figure*}

\noindent{\bf Depth stability over distance.} In order to evaluate the depth measurement stability of our method over difference distance, we conducted several case studies. We tested our method against a number of complex scene, one of those is shown in Fig.\ref{fig:figure6}. The same scene is observed by moving the ToF camera across 10 different viewing distance. Note that among these 10 captures the distance between the ToF camera and objects in the scene is distributed within 200cm. The data collected is consisted of ToF raw measurements captured under 200us, 400us and 4000us. Then, the collected ToF raw serve as the input of our trained network to generate depth map. The depth map generated under short exposure is compared with the one generated under 4000us. We observed that the variance of the error among each comparison is small and the average MAE between the depth map under 4000us exposure and the depth map under 200us/400us exposure is 4.8cm and 2.5cm respectively. This indicates both the precision and the accuracy of our method, when applied to very short exposure ToF raw, is comparable to a strong pipeline with 10 times longer exposure time. In many applications of ToF in consumer products, e.g. face recognition and photography in smartphones, the most widely used depth measuring distance is 30cm to 200cm. Thus the results in this experiment show the applicability of a very power efficient ToF design in this area.

\subsection{Qualitative results on our dataset}
\begin{figure}[!t]
\centering
\footnotesize
\renewcommand{\tabcolsep}{1pt} % adjust horizontal space
\renewcommand{\arraystretch}{1} % adjust vertical space
\begin{center}
\begin{tabular}{ccc}
  \includegraphics[width=0.32\linewidth]{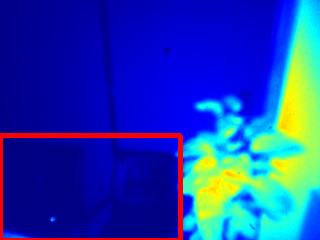} &
  \includegraphics[width=0.32\linewidth]{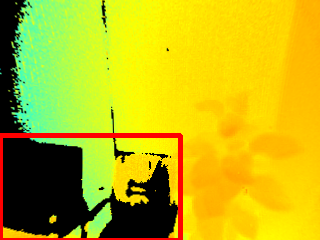} &
  \includegraphics[width=0.32\linewidth]{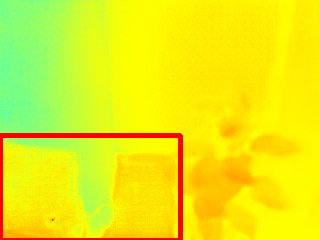}\\
  \includegraphics[width=0.32\linewidth]{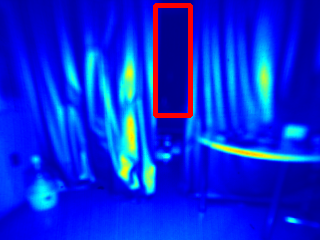} &
  \includegraphics[width=0.32\linewidth]{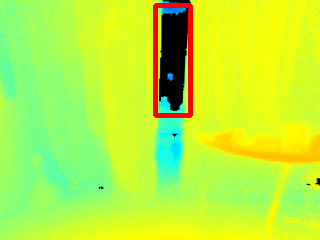} &
  \includegraphics[width=0.32\linewidth]{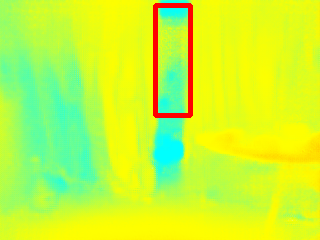}\\
(a) Amplitude&(b) ToF depth map with&(c) Our results from\\&suitable exposure time&ToF raw of (b)
\end{tabular}
\end{center}
\vspace{-2mm}
\caption{Traditional pipeline fails to recover the depth value of the regions marked out of the first group, since the black chair in the scene strongly absorbed signal emitted by ToF camera. For the regions marked out of the second group, too large distance results in few photons received by the ToF sensor. In contrast, our method is able to obtain high-quality depth maps for these two regions.}
\label{fig:figure7}
\vspace{-8mm}
\end{figure}
Then, we present the results of our method and the traditional ToF camera imaging pipeline in extreme cases on our test dataset. In this section, we verify that our proposed end-to-end solution can still reconstruct accurate depth value in extreme case. Moreover, compared with the traditional method, if the exposure time is set to regular, our method is more robust to scenes with objects of high absorptivity or regions in distance. In addition, we also explored multiple setting of exposure time or the active illumination power under which our method may fail. We note that other work \cite{Alpher23} also implements end-to-end imaging of ToF cameras, but their model training requires a large scale synthetic dataset, making it difficult to compare directly on our dataset.\\\vspace{-1mm}

\noindent{\bf Qualitative results with different exposure time.} We have shown that the amplitude value is an important indicator for evaluating the raw data quality of the ToF camera. Considering that the effect of the power level of the ToF camera active illumination system on the amplitude value in the amplitude map is equivalent to the effect of the length of exposure time, we simulate the power level of active illumination by controlling the length of the exposure time. In order to verify the effectiveness of our proposed method, we used the ToF raw measurements acquired under exposure time of 200us and 400us as the input of the network to predict the corresponding depth map. As shown in Fig.5, our results have better performance, compared with the depth map generated by the traditional ToF pipeline. Experiments show that our method is able to robustly process ToF camera raw with the exposure time of one order of magnitude shorter than that used in conventional ToF cameras. \\\vspace{-1mm}
 
\noindent{\bf Robustness under regular exposure.} Since there may exist some objects with low reflectivity or too large distance in the scene, choosing the appropriate exposure time or a strong power active illumination does not guarantee that the depth map of the entire scene is of high quality. However, our proposed method has better  performance in the depth estimation of these objects, compared with traditional ToF process, due to the ability of translating the weak and noisy ToF camera raw to depth map directly. As shown in Fig.\ref{fig:figure7}, we deliberately collected some scenes with dark objects and scenes with large distances(such as black stools and computer screens, glass doors with specular reflections, as well as objects with particularly large depth differences in the scene) to prove the robustness of our method in this case.\\\vspace{-1mm}

\begin{figure}[!t]
\centering
\footnotesize
\renewcommand{\tabcolsep}{1pt} % adjust horizontal space
\renewcommand{\arraystretch}{1} % adjust vertical space
\begin{center}
\begin{tabular}{cc}
  \includegraphics[width=0.45\linewidth]{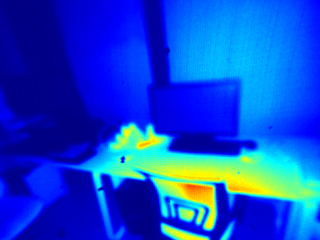} &
  \includegraphics[width=0.45\linewidth]{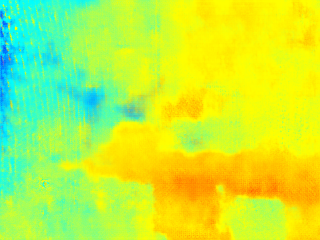} \\
  Amplitude graph & Results with 100us
\end{tabular}
\end{center}
\vspace{-2mm}
\caption{Failure case. Although the goal of our proposed method is to solve the depth mapping problem in low-power active illumination, our approach is likely to fail for extremely active illumination power, when we reduce the exposure time to 100us.}
\vspace{-4mm}
\label{fig:figure8}
\end{figure}

\noindent{\bf Failure cases.} Our proposed method aims to solve the depth mapping problem in low-power active illumination. However, this may fail if the illumination power is too low. For instance, when we reduce the exposure time to 100us, shown in Fig.\ref{fig:figure8}, the quality of the generated depth map is not satisfactory enough for some applications.
\section{Discussion and Conclusion}

\subsection{Implication to ToF camera design}
Using neural network to robustly process ToF camera raw with very short exposure time (raw data with low SNR) is a novel alternative to optimize the power efficiency of the whole ToF system. Despite the involvement of neural network computation, the inception of many recent low power neural network hardware makes it a practical solution. In addition to lowering the power consumption of ToF system, the results of this paper also provide a few extra design choices. First, higher depth frame rate may be achievable because the exposure time can be significantly reduced. Second, with the proposed method much smaller pixel size may be considered despite the SNR of the sensor raw could be low. Thus, higher depth resolution can thus be obtained with a reasonable power consumption. Such possibilities pave the way for new innovation in the ToF camera design.

\subsection{Concluding remarks}
In this paper, we discover that it is possible to devise a deep learning model to recover high quality depth information from very weak and noisy ToF raw measurements using deep learning. To realize the learning process, we collected a comprehensive dataset using a real-world ToF camera. We show in the experiments that our proposed method is able to robustly process ToF camera raw with the exposure time of one order of magnitude shorter than that used in conventional ToF cameras. While this neural network approach forms a key building block of a very power efficient ToF camera, it also shed new light on new innovations of the ToF camera design. We will make our code and dataset publicly available.

For future research, we will continue to improve the quality of our datasets. Specifically, we would adopt HDR imaging to improve the quality and precision of the ground truth depth map. Another opportunity for future work is to explicitly model the correction of the MPI error in an end-to-end trainable model to further enhance the accuracy of the results.

\clearpage

{\small
\bibliographystyle{ieee}
\bibliography{egbib}
}

\end{document}